# Evaluating Automatically Generated Phoneme Captions for Images

*Justin van der Hout*[1], *Zoltán D'Haese*[2], *Mark Hasegawa-Johnson*[3], *Odette Scharenborg*[1]

[1]Multimedia Computing Group, Delft University of Technology, Delft, The Netherlands
[2]KU Leuven, Leuven, Belgium
[3]University of Illinois, Urbana-Champaign, IL, USA

J.R.T.E.vanderhout@student.tudelft.nl, zoltan.dhaese@student.kuleuven.be, jhasegaw@illinois.edu,
O.E.Scharenborg@tudelft.nl

## Abstract

Image2Speech is the relatively new task of generating a spoken description of an image. This paper presents an investigation into the evaluation of this task. For this, first an Image2Speech system was implemented which generates image captions consisting of phoneme sequences. This system outperformed the original Image2Speech system on the Flickr8k corpus. Subsequently, these phoneme captions were converted into sentences of words. The captions were rated by human evaluators for their goodness of describing the image. Finally, several objective metric scores of the results were correlated with these human ratings. Although BLEU4 does not perfectly correlate with human ratings, it obtained the highest correlation among the investigated metrics, and is the best currently existing metric for the Image2Speech task. Current metrics are limited by the fact that they assume their input to be words. A more appropriate metric for the Image2Speech task should assume its input to be parts of words, i.e. phonemes, instead.

## 1. Introduction

Automatic image captioning [1], the generation of descriptions for images, is a popular task that combines the fields of computer vision and natural language processing (NLP). Image captioning systems typically use images with corresponding textual descriptions as training material. Unfortunately, such systems are only applicable to languages that have a conventional writing system (or well-defined orthographic system). Several languages around the world however do not have such an orthography [2]. In order to potentially reach any spoken language, regardless of whether it has an orthography, a new task has been proposed: Image2Speech [3], which takes an image as input and generates a caption as output. The main difference between Image2Speech and regular image captioning is that Image2Speech focuses on generating a *spoken* description without the use of textual descriptions. Rather than generating written words from image features, the Image2Speech system generates speech units (phonemes), which can then be synthesized into speech. Image2Speech circumvents the need for an orthography and it is therefore applicable to any spoken language.

Because the Image2Speech task is new, no established methods for evaluating the performance of a system for this task as yet exist. This paper aims to fill this gap. Specifically, we evaluated the output of the Image2Speech system, which consists of sequences of phonemes, with several objective metrics from the field of NLP. In order to determine how effective these metrics are, we correlate them with human ratings of how well the caption describes the image, collected via crowdsourcing. Since we are primarily interested in how well the phoneme sequences generated by the Image2Speech system describe the images, in other words, how well the semantics of the image are represented in the phoneme sequences, we focused on evaluating the semantics of the generated phoneme sequences rather than how well these phoneme sequences sound. For this purpose, 1) the phoneme sequences are converted into words so that they are readable and interpretable for the human raters; 2) we looked into more specific aspects of the images, namely objects and actions, to gain more insight into which aspects are most important to determine the goodness of the description of the caption.

## 2. Methodology

A new phoneme captioning system that is based on the Image2Speech system [3] has been developed, in order to obtain generated phoneme captions. An overview of the system can be found in Figure 1. Below we summarise the architecture and indicate where we deviate from the original system.

The system first extracts image features from the input with the VGG16 model (see Section 2.2.). These image features are then used as input for the image-to-phone model to generate a caption consisting of phonemes (see Section 2.3). The last step would be to use an audio synthesis model to synthesize a spoken caption from the phoneme sequences. This part is not yet implemented.

### 2.1. Data

The datasets that were used were the Flickr8k [4] image and text caption corpus and its associated Flickr-Audio corpus [5]. The Flickr8k corpus contains 8000 images from Flickr, with five textual captions for each of these images, totalling 40,000 captions. The Flickr-Audio corpus contains recordings of each of the 40,000 captions being read aloud. Both datasets were created with the use of Amazon Mechanical Turk workers. To train the Image2Speech system, phonetic transcriptions (in ARPABET) of the audio were used, which were created using the Janus Recognition Toolkit [6], and identical to those used in [3].

The training set consists of 6,000 images with their captions, while the validation and test sets each contain 1,000 images and their captions. However due to the automatic phonetic transcription sometimes failing, 5,956 images were used for training, 941 for validation and 959 for testing with up to 5 captions per image totalling 28,205 captions for testing, 4,741 for validation, and 4,705 for testing.

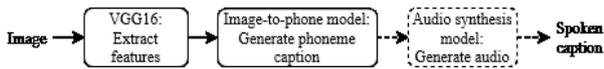

Figure 1: The Image2Speech system. Dotted parts are not yet implemented.

### 2.2. Image Features

In order to train a model that uses images as input and generates a sequence of phonemes describing the image as output, image features are required that can capture the most important elements of an image. We use VGG16 which is a convolutional neural network model developed by Simonyan and Zisserman [7]. This network, consisting of 13 convolutional layers and 2 fully connected layers, has been trained on ImageNet [8] which is a dataset that consists of over 14,000,000 images for roughly 22,000 nouns that come from WordNet [9]. In order to obtain image features from VGG16, the network has been cut off at the last convolutional layer. The size of this layer is $14 \times 14 \times 512$, or 196 sequential feature vectors of dimension 512, with every feature vector representing a $40 \times 40$ window of the original $224 \times 224$ image.

### 2.3. Image-to-phone model

XNMT (The eXtensible Neural Machine Translation Toolkit) [10] has been used to train the image-to-phone model. XNMT is a neural network-based toolkit that is specialized in machine translation and general sequence-to-sequence modelling. The image-to-phone model is a sequence-to-sequence model, which is why the image features are represented as a sequence.

The model is trained using the sequential image features as input and phonemic transcriptions of its captions as labels/output. While up to 5 captions per image are available, XNMT does not have an inbuilt functionality that can take multiple captions into consideration during training. Instead every image is paired up once with each of its captions, resulting in 5 training data points for an image that has 5 captions. The image-to-phone model has 3 main components: an encoder, an attender, and a decoder. The encoder uses a pyramidal LSTM (implemented with XNMT's PyramidalLSTMSeqTransducer) with 3 layers and a hidden dimension of 128. The attender uses a multi-layer perceptron (XNMT's MlpAttender) with a state dimension of 512 and a hidden dimension of 128. The decoder is implemented with XNMT's autoregressive-decoder which uses a one-directional LSTM with 3 layers and a hidden dimension of 512, and a multi-layer perceptron with a hidden dimension of 1024 as a layer of transformation between the LSTM and a final softmax layer. XNMT's default loss function was used which calculated the maximum likelihood loss. The main changes from the original system's architecture [3] are an increase of the encoder layers from 1 to 3 and an increase of the attender state dimension from 128 to 512.

### 2.4. Evaluation metrics

The original Image2Speech model was evaluated using BLEU4 scores and phoneme error rate (PER). In addition to these, we consider three other metrics which have all been developed for the evaluation of NLP-related tasks:

- **BLEU** [11] (bilingual evaluation understudy) is a popular metric for machine translation. It makes use of a modified precision, which is calculated by counting the number of n-grams in the hypothesis that can be matched in any of the references and dividing it by the total number of n-grams in the hypothesis. There are different BLEU scores depending on the highest order n-gram used. For example, BLEU4 uses n-grams up to $n = 4$ and is the most widely used BLEU metric. This metric was used to determine the performance of the image to speech system created by Hasegawa-Johnson et al. [3]. However they computed BLEU scores separately for every reference and averaged over all of them instead of computing a BLEU score for every candidate. To be fair in our comparison, we compute the BLEU score using both methods.
- **PER** [12] is calculated by summing each inserted, deleted or substituted phoneme in the output compared to the reference, divided by the total number of phonemes in the reference transcription.
- **CIDEr** [15] (Consensus-based Image Description Evaluation) is a metric intended to be used for automatic image captioning. It computes the similarity of a sentence with a set of multiple reference sentences. It uses Term Frequency Inverse Document Frequency (TF-IDF) to give lower weights to n-grams that appear frequently in the corpus. It then computes a CIDEr_n score which is the average cosine similarity between the hypothesis and the references for n-gram of size n. The CIDEr score is computed by taking the average of CIDEr_1 to CIDEr_4.
- **METEOR** [13] (Metric for Evaluation of Translation with Explicit ORdering) is another popular metric for machine translation. It creates an alignment between the candidate and reference sequences, by cutting the hypothesis into chunks. Then the harmonic mean of the precision and recall is calculated, with recall weighing more than the precision. The final score is computed by discounting the harmonic mean for the number of chunks that were required for the alignment.
- **ROUGE** [14] (Recall-Oriented Understudy for Gisting Evaluation) is a metric intended for machine translation and text summarization. It is similar to BLEU but as the name implies it is more focused on the recall, while BLEU is focused on the precision. This paper use ROUGE-L which is one of several variants of ROUGE and uses the Longest Common Subsequence (LCS). One advantage of using LCS is that there is no need to define an n-gram length, as it automatically uses the longest n-gram possible.

The objective metric scores are computed with nlgeval [16], except for PER[1].

### 2.5. Human evaluation

The effectiveness of these metrics is evaluated by correlating the metrics with human ratings of the captions generated by the Image2Speech system. The human ratings were collected with the use of Amazon's Mechanical Turk (MTurk), for monetary compensation. The Human Intelligence Tasks (HITs) for this experiment have been set up using the output of the iteration that performed best on the BLEU4 metric.

#### 2.5.1. Phonemes to words

Since crowdsource workers are typically not trained to read ARPABET phoneme sequences, the phoneme sequences are first converted into normal sentences of words. As already pointed out by [3], most of the generated phoneme sequences are interpretable as words. This conversion into words was done using a simple weighted finite state transducer (wFST). The wFST is a weighted graph with a circular path for every word where every node in the path represents a phoneme. The weights are exponentially lower for words with more phonemes, in order to prevent larger words that contain smaller words from being ignored. The wFST takes as input the generated phoneme sequence and a lexicon containing the phonetic transcriptions of the words used in the flickr8k dataset. The wFST computes the shortest path through the graph spanned by the lexicon and the generated phoneme sequences, and outputs the words on that shortest path. Because this method does not distinguish between similar sounding words, the output is manually corrected.

---

[1] https://holianh.github.io/portfolio/Cach-tinh-WER/

*2.5.2. Crowdsource evaluation*

The 952 test image/caption pairs were divided into 34 lists of 28 pairs without any overlapping pairs. Additionally, each list contained two control image/caption pairs with made-up captions: one image had a very bad caption and one image had a very good caption, which made for a total of 30 image/caption pairs per list. The control image/caption pairs were used to filter out raters who deviated too much from what was expected, e.g., due to a misunderstanding of the task. Every HIT contained one list of 30 image/caption pairs to be evaluated and every HIT was evaluated by five different evaluators.

We ran three separate experiments. Experiment 1 asked the participants to rate how well a caption described its corresponding image on a scale ranging from 1 (Very bad) to 7 (Very good). Experiments 2 and 3 asked the raters to rate how well the caption described the objects or actions, respectively, in the image on a scale from 1 (Very bad) to 4 (Very good). Prior to taking part in the HIT, the raters were provided with a number of example image/caption pairs from both ends of the rating scale to help them understand how to interpret the scale. Raters were able to evaluate multiple lists and participate in multiple experiment but could not rate a list that they had already evaluated. Raters were compensated with $0.60 for every HIT, which on an hourly basis is roughly equivalent to the minimum wage of Amazon workers.

## 3. Results

For 952 images of the test set, phoneme captions were successfully generated. Of the 952 captions, 921 captions make a proper sentence, i.e., are fully comprised of words, and contain a subject and verbs. After converting the phonemes into words, the total number of words that was generated (tokens) was 11,060 and the number of unique words (types) was 255, making for a type/token ratio of 0.023. The ground truth, i.e., the textual captions of the corpus has a type/token ratio of 0.020. The output of the Image2speech system thus shows good lexical diversity. Informal comparisons between different versions of the Image2Speech model showed that lexical diversity tended to increase with increasing BLEU4 score as computed by XNMT.

Table 1 shows the results of our implementation of Image2Speech and those of the original model of [3] using the metrics BLEU4 and PER metrics as computed by XNMT and reported in [3]. Please note that these metrics are computed slightly differently from those described in Section 2.4: XNMT does not take multiple references into account. The BLEU4 and PER scores are simply the average score over all image/caption pairs, rather than over all images. Table 1 shows that our model obtains a better BLEU4 score and a worse PER score compared to Hasegawa-Johnson et al.[3].

Table 1: *Image-to-phoneme model comparison with Hasegawa-Johnson et al. [3]. These scores have been computed using XNMT[10].*

| Metric | Our Model | Hasegawa-Johnson et al. [3] |
|---|---|---|
| BLEU4 | 15.6 | 13.7 |
| PER | 86.4 | 84.9 |

### 3.1. Human evaluation

The distribution of the results of the human evaluation regarding the overall quality of the captions can be found in Figure 2. The average overall score is 3.4 (±1.3), which would be between "Somewhat bad" and "Neutral". Figure 3 gives an example of a very good caption (left) and a bad caption (right). The results for the evaluations of actions and objects can be found in Figure 4. The average score for how well the actions are described by the captions is 2.1 (±0.8) and the average score for objects is 2.2 (±0.7), which in both cases roughly corresponds to "bad" on their scale. The difference in number of object and action ratings is due to more HITs being rejected for the latter.

Actions obtained a moderate to strong Pearson correlation of 0.57 ($p<0.001$) with the ratings of overall quality and objects obtained a moderate to strong correlation of 0.63 ($p<0.001$).

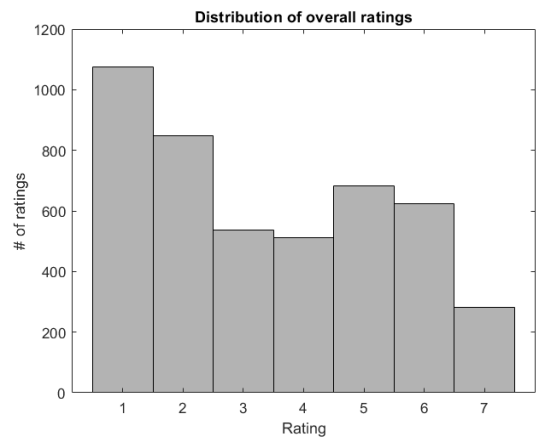

Figure 2: *Distribution of overall ratings obtained from Amazon's Mechanical Turk.*

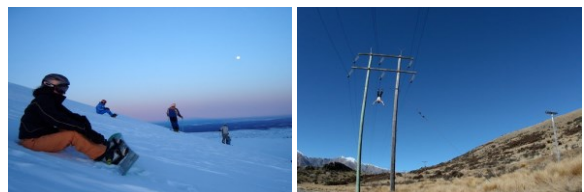

Figure 3: Examples of a very good and a bad caption.
*Left image (rated 6.4) captioned:*
"EY G R UW P AX F S K IY R Z AXR S K IY IX NG D AW N EY S N OW IY HH IH L" *("A group of skiers are skiing down a snowy hill.")*.

*Right image (rated 2.0) captioned:* "EY M AE N IH N EY Y EH L OW SH ER T IH Z S T AE N D IX NG AA N AX S T R IY T" *("A man in a yellow shirt is standing on a street.")*

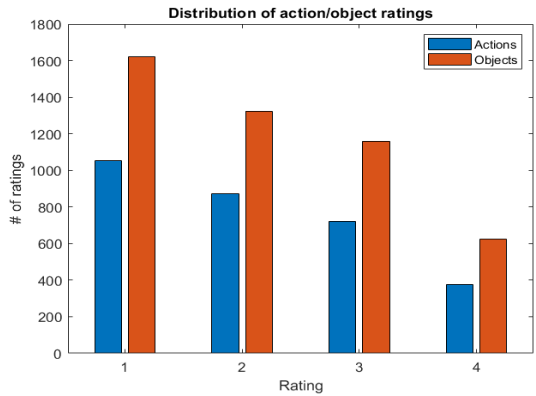

Figure 4: *Distribution of action and object ratings. There are more object ratings due to filtering.*

Table 2: *Average scores of the image-to-phoneme model over 5 iterations computed for different evaluation metrics (higher scores are better except for PER), and their correlations with the human ratings.*

| Metric | Score | $r$ | $r_{actions}$ | $r_{objects}$ |
|---|---|---|---|---|
| MTurk | 3.40 | | 0.569 | 0.627 |
| BLEU1 | 82.6 | 0.155 | 0.214 | 0.195 |
| BLEU2 | 61.3 | 0.355 | 0.388 | 0.411 |
| BLEU3 | 46.4 | 0.425 | 0.446 | 0.486 |
| BLEU4 | 36.1 | 0.435 | 0.449 | 0.494 |
| BLEU5 | 24.6 | 0.429 | 0.435 | 0.484 |
| BLEU6 | 18.2 | 0.410 | 0.406 | 0.451 |
| BLEU7 | 13.7 | 0.378 | 0.373 | 0.423 |
| BLEU8 | 9.3 | 0.340 | 0.319 | 0.376 |
| METEOR | 29.4 | 0.258 | 0.265 | 0.322 |
| ROUGE-L | 49.3 | 0.425 | 0.416 | 0.485 |
| CIDEr | 42.4 | 0.272 | 0.305 | 0.315 |
| PER | 71.4 | −0.361 | −0.363 | −0.381 |

### 3.2. Objective metrics

The results of the image-to-phoneme model in terms of the various objective metrics can be found in Table 2. We correlated the objective metrics with the three human ratings (overall, actions and objects). Table 2 shows the scores (higher scores are better except for PER) and Pearson correlation $r$ for every metric ($p<0.001$ for all metrics).

BLEU4 had the best correlation with the overall ratings, which corresponds to a weak to moderate correlation with the human ratings. BLEU5, BLEU3, ROUGE-L, and BLEU6 showed a weak to moderate correlation. BLEU7, BLEU2, BLEU8, PER, CIDEr, and METEOR only have a weak correlation. BLEU1 barely shows any correlation. The correlations for the action ratings are stronger for most metrics and even more so for the object ratings. In both cases, BLEU4 remains the strongest correlating metric.

## 4. Conclusion and Discussion

This paper presents an investigation on how to evaluate the relatively new task of generating descriptive speech units, or phonemes, from images, without the use of textual resources.

Human evaluation was obtained through Amazon's Mechanical Turk. Several Natural Language Generation metrics were then compared with the human ratings in order to establish which metric correlates best with human evaluation. The BLEU4 metric obtained the highest correlation with the human ratings, closely followed by BLEU5, BLEU3 and ROUGE-L. This pattern was also found when more specific aspects (i.e., actions and objects) were rated instead of the overall quality of the caption. Correlations between metrics and ratings of specific aspects were generally stronger than between metrics and ratings of overall quality. This may indicate that most metrics are better at evaluating these aspects rather than the overall quality, however it could also be caused by the fact that a 7-point scale was used for the overall ratings, and a 4-point scale for the ratings of actions and objects as this may have changed the behavior of the raters. Note, it is possible that the correlation results are biased towards the BLEU4 metric, however manual inspection of intermediate results indicated that increases in BLEU4 scores generally led to increases in lexical diversity and descriptive accuracy.

It is notable that the metrics with the highest correlation make use of medium length n-grams (i.e., 3-grams, 4-grams, 5-grams and LCS). CIDEr is an exception to this; it is possible that the use of TF-IDF has an adverse effect in this specific situation. Most generated sentences start in a very similar manner (e.g. "A dog", "A man" or "A group of people"). As a result, TF-IDF assigns a lower weight to these phrases, even though they are very important parts of the caption.

A new image-to-phoneme model has been trained on image/caption pairs from the Flickr8k database. Compared to the previous Image2Speech system, the new image-to-phoneme model obtained a better BLEU4 score. Adding more complexity to the model by increasing the hidden dimensions and the number of layers seemed to be beneficial for the task. Nevertheless, the human ratings showed there is still a lot of room for improvement.

Although BLEU4 obtained the highest correlation with human evaluation, it is not a strong correlation and is therefore not a perfect representation of human evaluation. At this moment however, BLEU4 is the best available indicator. Currently there is no metric that is specifically designed to determine the semantic similarity between an image and a sequence of phonemes. Such a metric could make use of n-grams of varying length, since words are made up of varying numbers of phonemes. It is important to note that phonemic unigrams usually do not have much semantic meaning, while higher order n-grams will often capture multiple words or half of two words. For that reason, it might be useful to try to identify which n-grams of phonemes correspond to a word. If that is possible to a sufficient degree, then currently existing metrics (See section 2.4) could be used the way that they were initially intended. Future research could also make use of other languages than English, particularly languages without an orthography for which the Image2Speech task is designed to test the assumption that this task can successfully be applied to unwritten languages.

## 5. Acknowledgements

The authors thank Markus Müller for creating the phonetic captions of the Flickr8k corpus, and the workers from Amazon Mechanical Turk for evaluating our captions.